\documentclass[letterpaper, 10 pt, conference]{ieeeconf}  %

\usepackage{times}
\usepackage{epsfig}
\usepackage{graphicx}
\usepackage{amsmath}
\usepackage{amssymb}
\usepackage{listliketab}
\usepackage{tabto}
\usepackage[dvipsnames]{xcolor}
\usepackage{float}

\usepackage{paralist}

\usepackage{amsmath}
\usepackage{amssymb}
\usepackage{mathtools,xparse}
\usepackage{algorithm}
\usepackage[noend]{algpseudocode}

\makeatletter
\def\BState{\State\hskip-\ALG@thistlm}
\makeatother

\IEEEoverridecommandlockouts                              %

\usepackage{paralist}

\newcommand{\timeHorizont}{N}

\newcommand{\vp}{\vec{p}}

\definecolor{brilliantrose}{rgb}{1.0, 0.33, 0.64}
\definecolor{amber}{rgb}{1.0, 0.49, 0.0}

\renewcommand{\vec}{\boldsymbol}

\newcommand{\mat}[1]{\boldsymbol{#1}}

\newcommand{\velzstate}{v_z} 						%

\newcommand{\thetastate}{\nu}

\newcommand{\thetavelstate}{\dot{\nu}}

\newcommand{\pospath}{\vec{s}}

\newcommand{\pospathprime}{\pospath^{\prime}}

\newcommand{\frenetfamevec}[1]{\vec{n}_{#1}}

\newcommand{\figref}[1]{Fig.\ \ref{#1}}

\newcommand{\R}{\mathbb{R}}

\title{\LARGE \bf Sample Efficient Learning of Path Following and Obstacle Avoidance Behavior for Quadrotors}

\author{Stefan Stev{\v{s}}i{\'{c}}$^{1}$, Tobias N{\"a}geli$^{1}$, Javier Alonso-Mora$^{2}$,  Otmar Hilliges$^{1}$%
\thanks{This work was supported by  in parts by the Swiss National Science Foundation (UFO 200021L\_153644) and NWO domain Applied Sciences. We  are grateful for their support.}
\thanks{$^1$AIT Lab, Department of Computer Science, ETH Zurich, 	8092 Zurich, Switzerland         {\tt\footnotesize \{stefan.stevsic | naegelit | otmar.hilliges\}@inf.ethz.ch}} 
\thanks{$^2$Cognitive Robotics, Delft University of Technology, 2628 CD Delft, Netherlands {\tt\footnotesize j.alonsomora@tudelft.nl}}%
}

\begin{document}

\maketitle
\thispagestyle{empty}
\pagestyle{empty}

\begin{abstract}\label{sec:abstract}
In this paper we propose an algorithm for the training of neural network control policies for quadrotors.
The learned control policy computes control commands directly from sensor inputs and is hence computationally efficient.
An imitation learning algorithm produces a policy that reproduces the behavior of a path following control algorithm with collision avoidance.
Due to the generalization ability of neural networks, the resulting policy performs local collision avoidance of unseen obstacles while following a global reference path. 
The algorithm uses a time-free model predictive path-following controller as a supervisor. 
The controller generates demonstrations by following few example paths. This enables an easy to implement learning algorithm that is robust to errors of the model used in the model predictive controller. 
The policy is trained on the real quadrotor, which requires collision-free exploration around the example path. 
An adapted version of the supervisor is used to enable exploration. Thus, the policy can be trained from a relatively small number of examples on the real quadrotor, making the training sample efficient.

\end{abstract}

\section{Introduction and Related Work}\label{sec:intro}

\begin{figure}[t]
\centering
\begin{tabular}{c}
\includegraphics[width=0.95\columnwidth]{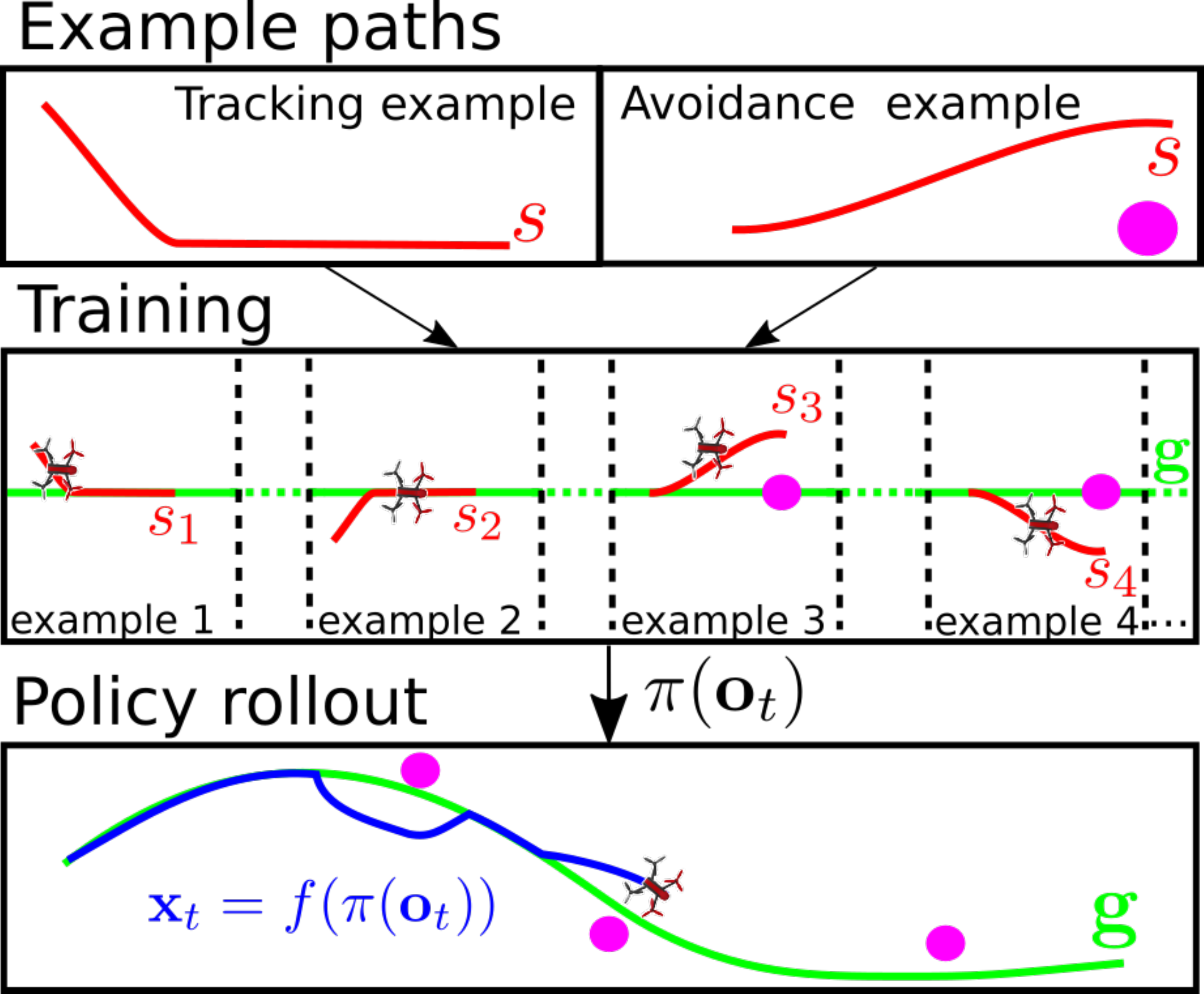}
\end{tabular}
\caption{\small{A policy is learned from few, short local collision avoidance and path following maneuvers (red). The learned policy generalizes to unseen scenes and can track long guidance paths (green) through complex environments while successfully avoiding obstacles (blue).}} \label{fig:setpointChange}
\end{figure}

Many applications of micro aerial vehicles (MAVs) require safe navigation in environments with obstacles and therefore methods for trajectory planning and real-time collision avoidance. Several strategies exist to make this problem computationally tractable. The use of model free controllers with path planning \cite{grzonka2012fully} is computationally attractive but requires conservative flight. Model based methods, including local receding horizon methods such as Model Predictive Control (MPC)~\cite{mueller2013model}, combining slow global planning with fast local avoidance \cite{oleynikova2016continuous}, or avoidance via search of a motion primitive library \cite{pivtoraiko2013incremental} are computationally demanding, but can achieve more aggressive maneuvers.
A theoretical analysis of the dynamical system can provide insights in a limited number of cases \cite{frew2004obstacle, rodriguez2014trajectory} leading to faster computation times. These methods have limited scope, taking into account a specific dynamics model. Furthermore, these methods require estimation of obstacle positions from the sensor data. In this paper, we address such issues with a novel imitation learning algorithm, schematically summarized in \figref{fig:setpointChange}, that produces control commands directly from sensor inputs.

Producing control signals directly form sensor inputs has two main benefits. First, the algorithm does not require estimation of obstacle positions. Second, function approximators, such as neural networks, can be much more computationally efficient compared to traditional planning methods \cite{mueller2013model, oleynikova2016continuous, pivtoraiko2013incremental} while still achieving safe flight. Learning can be combined with motion planning. Faust et al.~\cite{faust2014automated} combine learning of a low level controller with a path planning algorithm. This hybrid approach shows that control for quadrotor navigation can be learned, but still requires expensive off-line collision avoidance. Our method learns how to avoid collisions and runs in real time.  

The most general approach, to learn a controller, here called a control policy, is model-free reinforcement learning (RL) \cite{mnih2013playing}, a class of methods that learns the control policy through interaction with the environment. However, these methods are sample inefficient, requiring a large number of trials, and therefore can only be applied in simulation \cite{deisenroth2013survey}. A more sample efficient option is model-based RL \cite{abbeel2007application, levine2016end}, where the model parameters are learned while the control policy is optimized. In this setting, learning the model  requires dangerous maneuvers, which can lead to damage of the quadrotor or the environment \cite{KahnZLA16}. A final option is to learn the policy by imitating an oracle, either a human pilot or an optimization algorithm \cite{ross2011reduction}. If the oracle can provide examples of safe maneuvers, this is the most immediate choice to learn policies for real quadrotors.

Imitating the oracle is not a trivial task. Primarily because data from the ideal trajectory is not enough to learn a policy, since it does not provide examples of correcting drift from the ideal trajectory. In \cite{ross2013learning}, the policy learns to steer the quadrotor from a human pilot demonstration. The biggest challenge was to collect sufficient data, since it is challenging to control the quadrotor manually. As a result, the controls were limited to steering commands while speed and attitude were controlled externally. We resolve this issue by learning the policy from a trajectory optimization oracle. 

Imitating a trajectory optimization oracle requires to generate data that can efficiently train the control policy. Two main issues arise when training the policy with this approach. First, efficiently generating training data that can produce the control policy is not straightforward because the trajectory optimization is computationally demanding. Thus in prior work a single trajectory was used to compute control signals for states on the trajectory and states close to the trajectory \cite{mordatch2015interactive}. However, this approach only works in simulation with a perfectly known model. Second, the algorithm needs to work with an approximate model to be applied on a real world system. In \cite{zhang2016learning}, a long horizon trajectory was followed with a short horizon Model Predictive Controller (MPC) to generate training data efficiently. Since the model is not correct, this learning algorithm requires a complex adaptation strategy that guides the control policy to the desired behavior. Alternatively, one could use only a short horizon MPC to provide samples for training of the control policy \cite{KahnZLA16}. However, MPC can produce suboptimal solutions that can lead to deadlocks or collisions during training.

We present an algorithm which produces a control policy by learning from a Model Predictive Contouring Control (MPCC) \cite{lam2010model} oracle. Contrary to prior work \cite{zhang2016learning}, which relies on tracking of a \emph{timed} trajectory via MPC, MPCC is \emph{time-free}. More specifically:
\begin{inparaenum}[(i)]
  \item the policy shows faster execution time compared to non-learning approaches.
  \item MPCC allows for an easier to implement training algorithm that is robust to modeling errors. In comparison to the policy obtained by MPC, the MPCC based policy performs better and shows faster convergence behavior.
  \item A collision-free exploration strategy, bounding divergence from the collision-free region during on-policy learning.
\end{inparaenum}

The policy can be trained using paths of arbitrary length i.e. the planning horizon length is not a limit as in \cite{KahnZLA16}. As a result of (ii) and (iii), the algorithm is sample efficient, requiring a relatively low number of trials. This results in a training algorithm that can be executed on a real quadrotor.

\section{Preliminaries}\label{sec:preliminaries}

\newcommand{\path}{\pospath} %
\newcommand{\costK}{J_k} %
\newcommand{\errorCont}{e^c} %
\newcommand{\horLen}{N} %
\newcommand{\contProg}{\theta} %
\newcommand{\state}{\mathbf{x}} %
\newcommand{\inputVec}{\mathbf{u}} %
\newcommand{\policyOut}{\mathbf{y}}
\newcommand{\inputNN}{X} %
\newcommand{\outputNN}{U} %
\DeclarePairedDelimiter{\norm}{\lVert}{\rVert}
\newcommand{\observation}{\mathbf{o}}
\newcommand{\policy}{\boldsymbol{\pi}(\observation_t)} %
\newcommand{\dataset}{\mathcal{D}}
\newcommand{\datasetPath}{\mathcal{S}}
\newcommand{\vv}{\vec{v}}
\newcommand{\vg}{a_g}
\newcommand{\vn}{\vec{w}}
\newcommand{\vd}{\vec{d}}
\newcommand{\vl}{\vec{l}}
\newcommand{\globalPath}{\vec{\mathbf{g}}}
\newcommand{\vuPi}{\mathbf{u}_{\pi}} %
\newcommand{\vuS}{\mathbf{u}^{*}} %
\newcommand{\vuF}{\vec{u}}
\newcommand{\samplingT}{T_s}

\newcommand{\contourError}{\epsilon^c}
\newcommand{\lagError}{\epsilon^l}
\newcommand{\lagerrorApprox}{\hat{\epsilon}^l}
\newcommand{\contourerrorApprox}{\hat{\epsilon}^c}
\newcommand{\relPath}{\vec{r_p}}
\newcommand{\contourCost}{c^c}
\newcommand{\pathAprox}{\vec{\hat{s}}}
\newcommand{\gainSafe}{k_{safe}}

\newcommand{\rollstate}{\phi_d} 				%
\newcommand{\pitchstate}{\theta_d}				%
\newcommand{\delstatesroll}{\phi} 				%
\newcommand{\delstatespitch}{\theta}	        %
\newcommand{\dotdelstatesroll}{\dot{\phi}} 	    %
\newcommand{\dotdelstatespitch}{\dot{\theta}}	%
\newcommand{\quadyaw}{\psi}						%
\newcommand{\dotquadyaw}{\dot{\psi}}			%
\newcommand{\dotquadyawdes}{{\dot{\psi}}_{d}}			%

\subsection{Robot Model}\label{sec:robot_model}

The full state of a quadrotor is 12-dimensional, consisting of quadortor position, velocity, rotation and angular velocity. However, in our experiments we use a 8-dimensional state \cite{NaegeliSIG}, based on the Parrot Bebop2 SDK inputs. The state is defined by the position $\vp \in \R^3$, velocity $\vv \in \R^2$ in $x,y$ directions, and roll $\delstatesroll$, pitch $\delstatespitch$ and yaw $\quadyaw$:
\begin{align}
\state = [\vp,\vv, \delstatesroll, \delstatespitch, \quadyaw]\in \R^{8}.
\end{align}
The set of feasible states is denoted by $\mathcal{X}$. The control inputs to the system are given by $\inputVec = [\velzstate,\rollstate,\pitchstate, \dotquadyawdes]\in \R^4$, where $\velzstate$ is the velocity of the quadrotor in $z$ direction, $\rollstate$ and $\pitchstate$ are the desired roll and pitch angles of the quadrotor. The rotational velocity around the z-body axis is set by $\dotquadyawdes$. The set of feasible inputs is denoted by $\mathcal{U}$. Sets $\mathcal{U}$ and $\mathcal{X}$ are described in Sec. \ref{sec:mpcc-single}. We use a first order low-pass Euler
approximation of the quadrotor dynamics. Notice that velocities $\velzstate$ and $\dotquadyawdes$ are directly controlled. The velocity $\velzstate$ directly controls the position dynamics in $z$ direction $\dot{\vp} = [\vv, \velzstate]$. The dynamics of the state velocity vector are:
\begin{align}\label{eq:model-position}
\dot{\vv} = \mat{R}(\quadyaw)
\begin{bmatrix}
-tan(\delstatesroll)\\
tan(\delstatespitch)
\end{bmatrix}\vg - c_d \vv,
\end{align}
where $\vg = 9.81\frac{m}{s^2}$ is the earth's gravity, $\mat{R}(\quadyaw)\in SO(2)$ is the rotation matrix only containing the yaw rotation of the quadrotor and $c_d$ is the drag coefficient at low speeds.
The rotational dynamics of the quadrotor are given by
\begin{align}\label{eq:model-position2}
\dotdelstatesroll = \frac{1}{\tau_a}(\rollstate - \delstatesroll), \,\,\,
\dotdelstatespitch = \frac{1}{\tau_a}(\pitchstate - \delstatespitch) \,\,\,
\textnormal{and} \,\,\,
\dotquadyaw = \dotquadyawdes,
\end{align}
where $\tau_a$ is the time constant of a low-pass filter. As a result, the position $\vp$  and velocity $\vv$ cannot change instantaneously.

\subsection{Control policy}

Our work is concerned with dynamical systems, such as quadrotors, described by a state vector $\state$ and controlled via an input vector $\inputVec$ (Fig.~\ref{fig:physics_contour}). We assume that the system has sensors, such as a laser range finder, odometry etc. We denote sensor readings at time $t$ with an observation vector $\observation_t$.

Dynamical systems are typically controlled via a manually tuned control law e.g. PID or LQR control. In contrast, we approximate the control law by learning a control $policy$ from examples. The policy $\policy$ is a function which, at every time step $t$, takes the vector $\observation_t$ as input and outputs the system control inputs $\vuPi = \policy$. The control inputs  $\vuPi$ are independent of the time step, producing the same control vector for the same observation, i.e. the control policy is stationary and deterministic. 

We learn a policy for local collision avoidance, while following a global guidance $\globalPath$, coarsely describing the desired robot trajectory (cf. \figref{fig:setpointChange}, bottom). In our case, the guidance $\globalPath$ is a natural cubic spline. The guidance may be computed off-line based on mission goals or given by human and does not need to be collision-free. For example, maps provide information about walls or buildings, while obstacles like trees or humans are not represented. Thus, a control law that locally and at run time alternates the prescribed path, while continuing to follow the global mission goal is necessary.

\subsection{Policy inputs and outputs}\label{sec:pol_in_out}

\begin{figure}
\centering

\begin{minipage}{.60\columnwidth}
  \centering
  \includegraphics[width=.75\columnwidth]{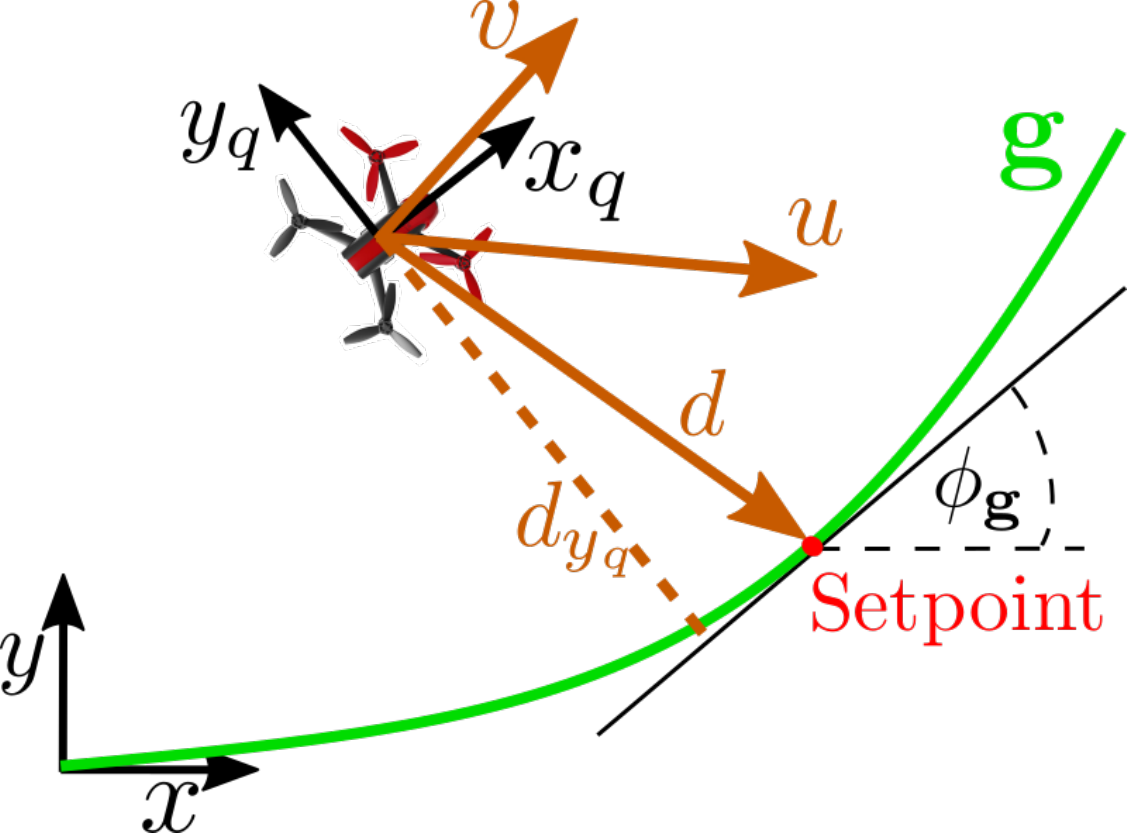}
\end{minipage}%
\begin{minipage}{.4\columnwidth}
  \centering
  \includegraphics[width=.75\columnwidth]{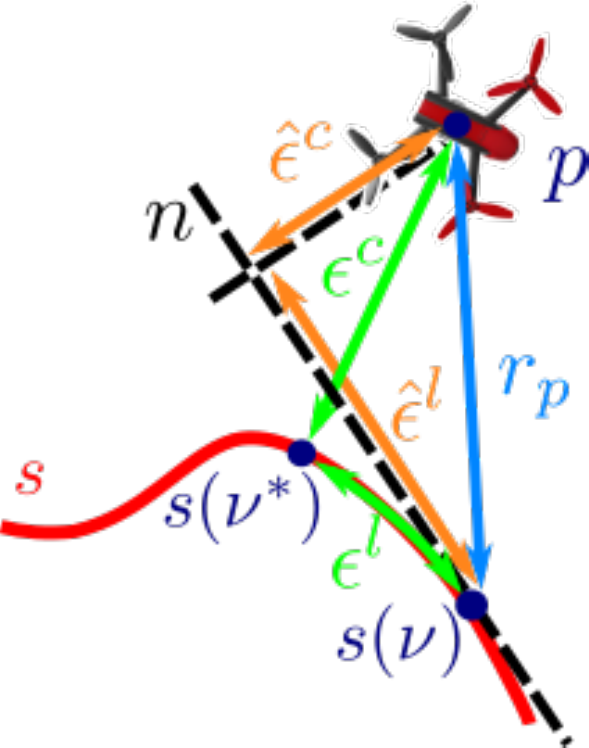}
\end{minipage}
\caption{\small{Left: \textbf{Coordinate systems:} Global and quadrotor coordinate systems. The quadrotor coordinate system is denoted with a subscript $q$. Policy inputs and outputs are always calculated in the quadrotor frame. Right: \textbf{Contouring error approximation:} Illustration of the real contouring and lag errors (green) as well as the approximations (orange) used in our MPCC implementation.}}
\label{fig:physics_contour}
\end{figure}

The input to the policy $\policy$ is an observation vector $\observation_t = [\vd_t, \vv_t, \vl_t]$ (Fig.~\ref{fig:train_test}), consisting of the distance to the guidance $\vd_t$ in the quadrotor $yz$-plane, the quadrotor velocity $\vv_t$ and laser range finder readings $\vl_t\in \R^{40}$. Distance measurements $\vd_t$ are obtained by subtracting quadrotor positions $\vp$ from the current setpoint on the guidance $\globalPath$ (\figref{fig:physics_contour}):
\begin{equation}\label{eq:clalc_d}
\vd_t = (\vp - \vp_{d}) R(\phi_{\mathbf{g}}) ,
\end{equation}
where $R(\phi_{\mathbf{g}})$ is a rotation matrix around $z$. The angle $\phi_{\mathbf{g}}$ is calculated from the global path tangent. The setpoint $\vp_{d}$ moves along the path with constant velocity. The distance in $x$ can be omitted since the quadrotor is trained to progress along the guidance $\globalPath$. Quadrotor positions $\vp$ and velocity measurements $\vv_t$ are obtained directly from the on-board odometry. The policy $\policy$ outputs continuous control signals: vertical velocity, and roll and pitch angles of the quadrotor $\vuPi = [\velzstate,\rollstate,\pitchstate]$. The heading is controlled separately (Sec.~\ref{sec:global_path}). This enables a simpler learning algorithm compared to learning of the full quadrotor input $\inputVec$. Generally speaking, $\observation_t$ could consist of arbitrary sensor data, such as depth images or ultrasound sensor readings.

\subsubsection{Sensor models} \label{sec:sensor_model}
At training time we have access to the full state $\state_t$. To obtained simulated observations $\observation^s_t= [\vd^s_t, \vv^s_t, \vl^s_t]$ we calculate $\vd^s_t$ via Eq. (\ref{eq:clalc_d}), where $\vp$ is taken directly from $\state_t$. Analogously, $\vv^s_t$ is taken from $\state_t$. The laser range finder readings are obtained by casting rays from the quadrotor position in the directions of the scanning laser $\vl^s_t = \boldsymbol{f}_l(\state_t, \vp_{ob})$, where $\vp_{ob}$ are obstacles position known at training time. We do not add any noise.

\subsection{Trajectory-tracking (MPC) vs Path-following (MPCC)}
Receding horizon tracking is the most common method (e.g., MPC used in \cite{zhang2016learning}) to steer a quadrotor along a trajectory. The trajectory is a sequence of state vectors with associated timings $([\state_i, t_i])^N_{i=1}$. The aim is to position the robot on a \emph{time-parametrized} reference trajectory i.e. to be at a particular position at each time step. The MPC optimization depends on the current time step and the quadrotor states.

We use a \emph{time-free} path-following objective in an MPCC formulation. In path-following control, the robot inputs are optimized to stay close to the desired path $\path$ and to make progress along the path \cite{lam2010model}. The desired path $\path$ is a geometric representation of desired robot positions $\vp$ during movement. Through the paper, we use $\path$ for paths followed by MPCC, while paths followed by the policy $\policy$ are refereed as global guidance $\globalPath$.  Control inputs $\inputVec$ are obtained from an optimization problem, which minimizes the distance to the desired path $\path$, and maximizes the progress along $\path$. The distance to the closest point on $\path$ is denoted by the contouring error $\contourError$:
\begin{equation}\label{eq:contEr}
\contourError = \| \path(\thetastate^*) - \vp \|,
\end{equation}
where $\path$ is a cubic spline parametrized with $\thetastate$ (cf. \figref{fig:physics_contour}, right). Finding the closest point on the path $\path(\thetastate^*)$ is an optimization problem itself and cannot be solved analytically. We discuss a computationally tractable solution in Sec.~\ref{sec:mpcc}.

\begin{figure}[t]
	\centering
	\includegraphics[scale=0.24]{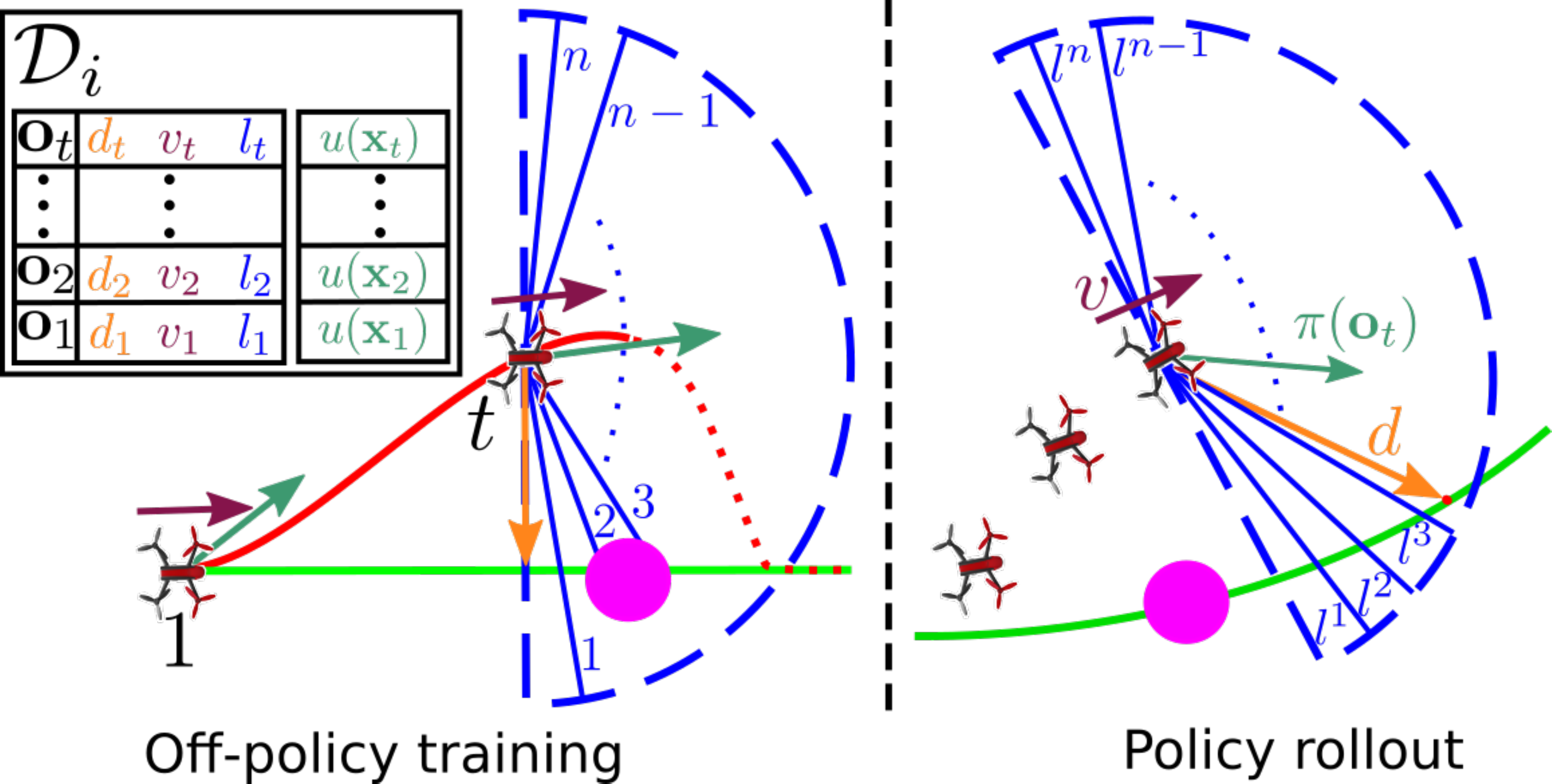}
	\caption{\small{Left: \textbf{Off-policy training example.} Observation data is collected during training while following the example path $\path_i$ with the MPCC controller. Right: \textbf{Global path following.} The policy produces control inputs based on the current observation vector.}}\label{fig:train_test}
\end{figure}

\section{Method}

\subsection{Policy Learning Algorithm}

We propose an imitation learning algorithm to iteratively refine a control policy $\policy$, learning a general behavior from a set $\datasetPath = \{\path_i\}^N_{i=1}$ of \emph{short} example paths $\path_i$.

The goal of the control policy $\policy$ is to imitate the trajectory produced by the MPCC supervisor, tracking the path $\path_i$. We employ supervised learning on the dataset of observation-control mappings $(\observation_t,\vuF (\state_t))$. The training data is obtained in a two-step procedure. First, an \emph{off-policy} step generates training samples via tracking the example path $\path_i$ with the MPCC oracle (cf. Fig.~\ref{fig:train_test}). However, this only produces ``ground truth'' data, containing samples from the ideal trajectory, which are not enough to train the control policy as observed in DAgger \cite{ross2011reduction}. We gather the necessary additional data by using the partially trained policy in an \emph{on-policy} step. Inevitably the policy outputs $ \vuPi = \policy$ will lead to drift from the ideal trajectory. Correct control inputs $\vuS = \vuF(\state_t)$, corresponding to recorded observations $\observation_t$, are computed by the MPCC supervisor after the data collection. 

\begin{figure*}[ht!]
\centering
	\includegraphics[scale=0.18]{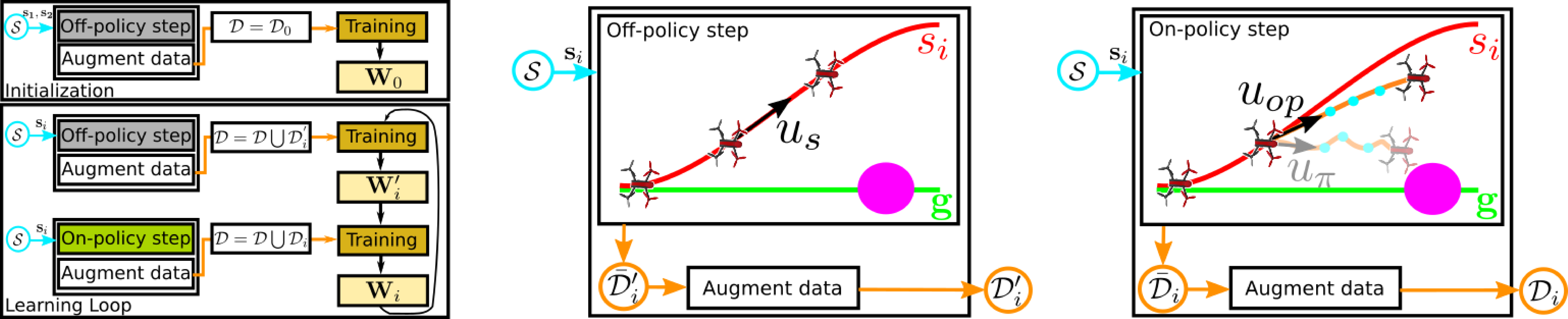}
	\caption{\small{\textbf{Overview:} The algorithm for training the policy $\policy$ (left). \textit{Off-policy} and \textit{on-policy} steps for data collection (middle and right).}}
	\label{fig:algorithm_details}
\end{figure*}

\subsubsection{Example paths}\label{sec:mot_prim}

We provide examples via simple heuristics (Fig.~\ref{fig:setpointChange}), demonstrating returning to the spline at $45^o$ and showing obstacle avoidance maneuvers starting $3~m$ from the obstacle and passing it at a distance of $1.5~m$. Each skill requires several examples of the same type. Importantly, these paths do not need to take the model of the robot into account. We use 12 example paths in total. The obstacles are cylinders of radius $r=0.2~m.$

\subsubsection{Policy representation}

The policy is parametrized by a universal function approximator in the form of a neural network. The network parameters define a matrix $\mathbf{W}$. The full notation is $\boldsymbol{\pi}(\observation_t; \mathbf{W})$, but we often use $\policy$ for brevity. We use a fully connected network with two hidden layers, each consisting of 30 neurons with softplus activation and linear neurons in the output layer. Initial weights $\mathbf{W}$ are initialized randomly using zero mean normal distribution with standard deviation $0.01$.

\subsubsection{Data collection}
To collect data for training, we have two different steps for which we use two different controllers: \textit{off-policy step} (MPCC) and \textit{on-policy step} (on-policy MPCC). Each of the steps is used to collect training data. The quadrotor tracks the given path $\path_i$ using the respective controller and we collect the observation samples $\observation_t$ and system state $\state_t$ at each time step.
 
\subsubsection*{Off-policy step (MPCC path tracking)}\label{sec:off_policy_alg}

In this learning step, ``ground truth'' training samples are collected while the quadrotor tracks the given path $\path_i$ using the MPCC supervisor. After the off-policy step the dataset $\dataset$ contains only ideal trajectory data.

\subsubsection*{On-policy step (on-policy MPCC path tracking)}\label{sec:on_policy_alg}
 
We propose an exploration approach to visit the states $\state_t$ that the non-fully trained policy $\policy$ would visit. For exploration, we use an \textit{on-policy MPCC} that generates control inputs $\inputVec_{op}$ (cf. Sec.~\ref{sec:on-policy}). The on-policy MPCC optimization cost balances between following the current policy output $ \vuPi = \policy$ and minimizing the contouring error, pulling the quadrotor back to the path (cf. Fig.~\ref{fig:algorithm_details}, right). This enables the collision-free exploration. Here, we assume that the region around the example path is safe and obstacle-free.

\subsubsection{Data augmentation and training}
\label{sec:data_augmentation}
The training dataset is constructed from collected observations $\observation_t$ and states $\state_t$. For each state $\state_t$, the MPCC supervisor computes the optimal trajectory and control inputs in the horizon, with respect to the path $\path_i$. However, only the first control input $\vuS = \vuF(\state_t)$ is used as a training sample: 
\begin{equation}
\bar{\dataset}_i = \{(\observation_t,\vuF(\state_t)), t = 1..n\}.
\end{equation}

We add noisy samples to the dataset $\bar{\dataset}_i$ to prevent overfitting during training \cite{mordatch2015interactive}. First, Gaussian noise is added to every state $\state_t$ collected during path tracking. The resulting noisy states $\state_t + \vn_{t}$ are used to calculate corresponding input samples $\vuF(\state_t + \vn_{t})$ via the MPCC supervisor. Observation samples $\observation^s(\state_t + \vn_{t})$ are obtained by calculating the exact observations from the noisy states using sensor models (cf. Sec. \ref{sec:sensor_model}). For each real sample, we add three noisy samples to augment the dataset $\bar{\dataset}_i$:
\begin{equation}
\begin{split}
\dataset_i = \bar{\dataset}_i \bigcup  \{(\observation^s(\state_t + \vn_{tk}),\vuF(\state_t + \vn_{tk})), \\ t = 1..n, k=1..3\}.
\end{split}
\end{equation}
We add the augmented dataset $\dataset_i$ to the global dataset $\dataset = \dataset \bigcup \dataset_i$. Using the new dataset $\dataset$, the policy $\policy$ is trained via optimizing the mean squared error (MSE) on $\dataset$:
\begin{equation}\label{eq:MSE}
\underset{\mathbf{W}}{\min} \sum_{\observation_j, \vuS_{j} \in \dataset} \norm{\boldsymbol{\pi}(\observation_j;\mathbf{W}) - \vuS_{j}}_2^2.
\end{equation}
The neural network is trained incrementally by initializing the network weights $\mathbf{W}$ from the previous solution. We use the ADAM optimization algorithm for training.

\subsubsection{Algorithm}
The algorithm (Fig.~\ref{fig:algorithm_details}) requires only a set of example paths $\datasetPath$ as input. Data collection is done on the real quadrotor because the on-policy data depends on the error of the approximate model. These are the most important steps:
\begin{itemize}
\item \textbf{Initialization}. We execute two \emph{off-policy} data collection steps on two return-to-guidance paths selected at random. The data is augmented (see Sec.~\ref{sec:data_augmentation}) and the initial policy is trained. The initial policy needs enough data to ensure stable performance in the \emph{on-policy} step.
\item \textbf{Learning loop}. During training the algorithm alternates between \emph{off-policy} and \emph{on-policy} data collection steps, augmenting data, and re-training the policy after every step using the remaining samples $\datasetPath \setminus \{\path_1, \path_2 \}$. The \emph{off-policy} step collects ground truth data. 
The \emph{on-policy} step helps to correct the behavior of the intermediate policy. 
\item \textbf{Output.} The final policy is trained from different examples. This enables the policy to generalize to different obstacle positions beyond the ones in the training set.
\end{itemize}

\subsection{Policy supervisor (MPCC)}\label{sec:mpcc}

To follow the path we seek to minimize the contouring error $\contourError$ defined in Eq. (\ref{eq:contEr}) and maximize the progress along the same path $\path$. To solve this problem we follow the formulation of \cite{NaegeliSIG}. We introduce an initial guess $\path(\thetastate)$ of the closest point $\path(\thetastate^*)$, which is found by solving the MPCC problem Eq.~(\ref{eq:mpcc-single}). The integral over the path segment between the closest point $\path(\thetastate^*)$ and location of $\path(\thetastate)$ denotes the lag error $\lagError$. To attain a tractable formulation, the errors $\lagError$ and $\contourError$ are approximated by projecting the current quadrotor position $\vp$ onto the tangent vector $\frenetfamevec{}$, with origin at the current path position $\path(\thetastate)$ (\figref{fig:physics_contour}, right). The relative vector between $\vp$ and the tangent point $\path (\thetastate)$ can be written as $\relPath: = \path (\thetastate) -\vp$. Using the path derivative $\pospathprime:=\frac{\partial \path(\thetastate)}{\partial\thetastate}$, the normalized tangent vector $\frenetfamevec{} = \frac{\pospathprime}{\|\pospathprime\|}$ is found. The approximated error measures are then given by:
\begin{subequations}
\begin{align}
\lagerrorApprox = \|\relPath^T\frenetfamevec{}\| ,
\end{align}
\begin{equation}
\contourerrorApprox =
\|\relPath-\left(\relPath^T\frenetfamevec{}\right)\frenetfamevec{}\| ,
\end{equation}
\end{subequations}

With these error measures, we define a stage cost function:
\begin{align}\label{eq:cost-path}
J_k = K_c{(\hat{\epsilon}^c_{k}})^2 + K_l ({\hat{\epsilon}^l_{k}})^2 -\beta \thetavelstate_k ,
\end{align}
where the subscript $k$ indicates the horizon stage in Eq.~\ref{eq:mpcc-single}. $J_k$ represents the trade-off between path following accuracy and progress along the path, where $\thetavelstate$ is a velocity of the parameter $\thetastate$, describing path progress and $\beta \geq 0$ is a scalar weight. The scalar weight $K_l$ determines the importance of the lag error and is set to a high value which gives better approximations of the closest point $s(\thetastate_k)$. The admissible contour error is controlled by the weight $K_c$.

\subsubsection{MPCC Formulation}\label{sec:mpcc-single}
The trajectory and control inputs of the drone at each time step are computed via solving the following $N$-step finite horizon constrained optimization problem at time instant $t$:
\begin{align}
\label{eq:mpcc-single}
\underset{\inputVec,\state,\thetastate,\thetavelstate}{\text{minimize}}\quad & \sum_{k=0}^{\timeHorizont} J_k + {\inputVec}_k^T \mat{R} {\inputVec}_k \\
\text{subject to}\quad &  \state_{k=0} = \state_t && \text{(Initial state)} \nonumber\\
& \thetastate_{k=0} = \thetastate_t && \text{(Initial path parameter)}  \nonumber\\
& \state_{k+1} = \boldsymbol{f}_m(\state_{k}, \inputVec_{k}) && \text{(Robot dynamics)} \nonumber \\
& {\thetastate}_{k+1} = {\thetastate}_k + {\thetavelstate}_k T_s && \text{(Progress path)} \nonumber \\
&  0 \leq \thetastate_k \leq l_{path} && \text{(Path length)} \nonumber \\
&  {\state}_k \in \mathcal{X}, && \text{(State constraints)} \nonumber \\
& \inputVec_k \in \mathcal{U}, && \text{(Input constraints)} \nonumber \\
\end{align}
where $\mat{R}$ is a positive definite penalty matrix avoiding excessive use of the control inputs. The vector $\state_t$ and the scalar $\thetastate_t$ denote the values of the current states $\state$ and $\thetastate$, respectively. The scalar $T_s$ is the sampling time. The state constraints $\mathcal{X}$ limit roll and pitch angles $ \delstatesroll, \delstatespitch$ to prevent the quadrotor from flipping. The input constraints $\mathcal{U}$ are set according to the quadrotor's allowed inputs. This non-linear problem under constraints~\eqref{eq:mpcc-single} can be formulated in standard software, e.g. FORCES Pro~\cite{forcesproweb}, where efficient code can be generated for real-time solving.

\subsection{Exploration algorithm (On-policy MPCC)}\label{sec:on-policy}

For \emph{on-policy} learning we apply a variant of the above MPCC, attained by adding a following cost to Eq. \eqref{eq:cost-path}:
\begin{equation}
\label{eq:safe_cost}
\begin{aligned}
c_k = \norm{\state_{\pi k} - \state_k}_2^2  .
\end{aligned}
\end{equation}
This term trades-off visiting states $\state_{\pi k}$ obtained by rolling-out the policy $\policy$, while keeping the quadrotor close to the input path $\path_i$. The main difference compared to the \emph{off-policy} supervisor is a larger admissible contouring error $\contourerrorApprox$.
In simulation  the policy $\policy$ is rolled-out over the entire horizon length to obtain the predicted quadrotor state $\state_{\pi k}$. The observation vector $\observation_k$ is computed from these states using the sensor models (cf. Sec. \ref{sec:sensor_model}).

The cost used in the on-policy MPCC is similar to the one presented in PLATO \cite{KahnZLA16}, where the quadrotor tries to greedily follow the policy output in the first state, while keeping the standard objective in the next states. We build on top of this cost to improve safety during the exploration. The \emph{on-policy} MPCC observes all policy states in the horizon $\state_{\pi k}$ which provides more complete information about the states. Furthermore, the exploration area is more precisely defined because the contouring cost is directly proportional to the distance from the collision-free example path.

\section{Method Discussion}

\subsection{Generalization and Limitations}

\begin{figure}[t]
	\centering
	\includegraphics[width=0.95\columnwidth]{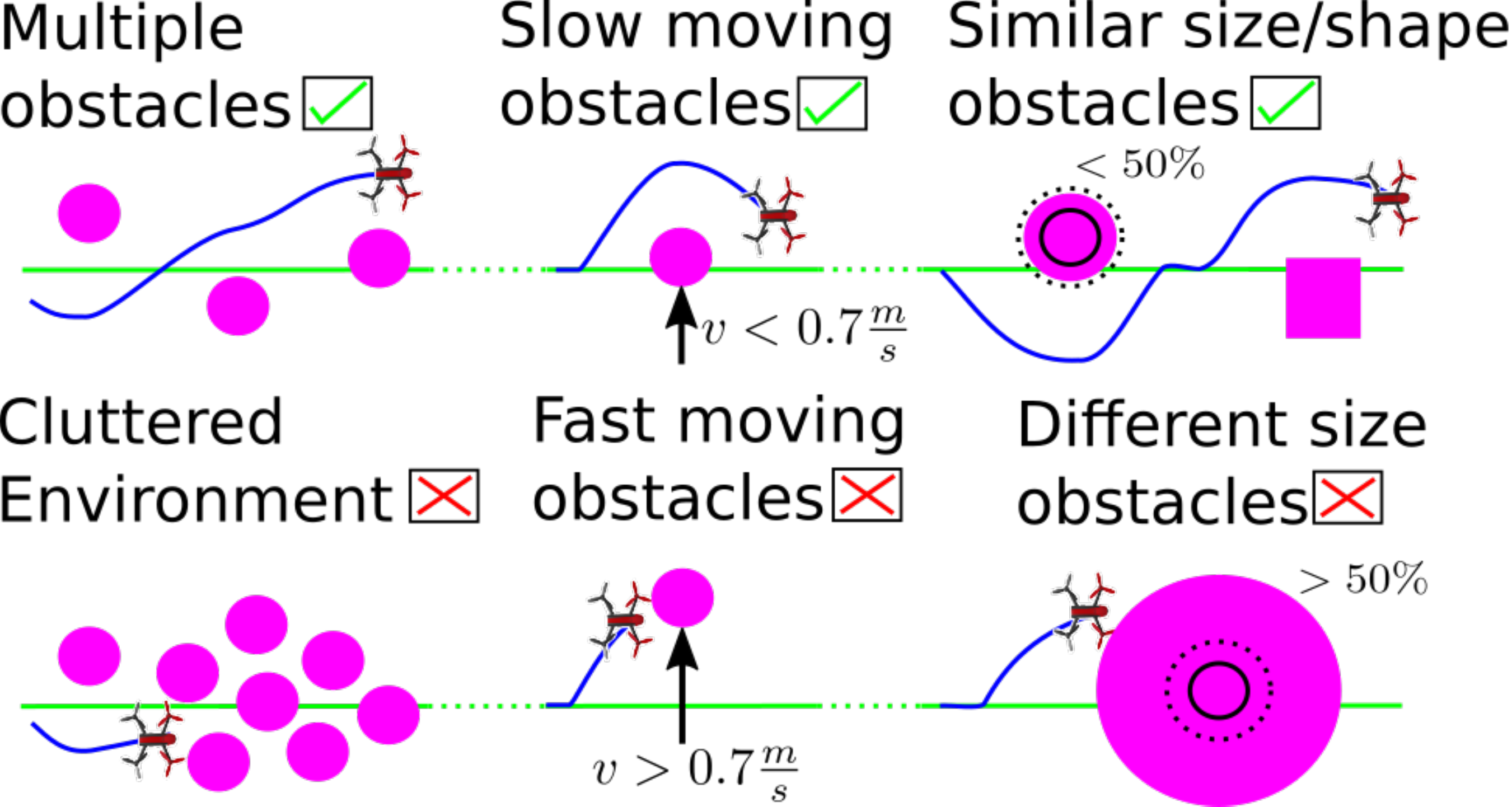}
	\caption{\small{\textbf{Generalization \& limitations.} Schematic of settings the policy generalizes to and limits of generalization. We experimentally verified that obstacles moving up to $0.7 \frac{m}{s}$ perpendicular to the quadrotor direction can be successfully avoided, while faster moving obstacles cannot. Changing the obstacle diameter up to $50\%$ compared to training, results in satisfying behavior. Further, different shaped obstacles of similar size can be avoided.}}\label{fig:generalisation}
\end{figure}

It is important to note that our learning algorithm never sees entire trajectories. Instead we provide multiple, short examples of a class of behavior. They provide guidance on how to react in different \emph{instances} of the same problem. The final policy $\policy$ generalizes to unseen scenarios (cf. Fig.~\ref{fig:generalisation}), following paths much longer than those seen during training.

These generalization properties can be explained from a machine learning perspective. Neural networks are universal function approximators, able to learn a function from a set of in- and output pairs. In our case, we assume that samples come from a non-linear stochastic function
\begin{equation*}
\vuS = \boldsymbol{f}_{nn}(\observation_t) + \boldsymbol{\varepsilon},
\end{equation*}
where $\boldsymbol{\varepsilon}$ is zero mean Gaussian noise $\mathcal{N}(\boldsymbol{0},~\boldsymbol{\sigma})$. The control inputs $\vuS$ directly depend on the system state $\state_t$, but we assume partial observability of the state $\state_t$ from the observation $\observation_t$. The function output $\vuS$ can be described by the conditional probability distribution
\begin{equation*}
p\,(\vuS~|~O = \observation_t; \mathbf{W}) = \mathcal{N}(\boldsymbol{\mu},~\boldsymbol{\sigma}),
\end{equation*}
where the distribution mean $\boldsymbol{\mu} = \boldsymbol{f}_{nn}(\observation_t)$ is parametrized by the neural network. Given the sample pair $(\observation_j, \vuS_j)$, for fixed network parameters $\mathbf{W}$, we can calculate the probability $P(U = \vuS_j~|~O = \observation_j; \mathbf{W})$. Maximum log likelihood estimation (MLE) yields the neural network parameters $\mathbf{W}$:
\begin{equation*}
\mathbf{W} = \underset{\mathbf{W}}{\text{arg\,max}} \sum_{\observation_j, \vuS_j \in \dataset}{ln~P(U = \vuS_j| O = \observation_j; \mathbf{W})}.
\end{equation*}
Since a Gaussian distribution is assumed, the mean can be obtained directly via the MSE loss in Eq. (\ref{eq:MSE}).

The policy $\policy$ is trained sequentially on example paths to achieve sample-efficient learning. However, the policy $\policy$ directly depends on the statistics obtained from the training samples in the final dataset $\dataset$. The MLE principle provably provides the best fit to the given samples, which leads to good generalization properties in cases where test samples come from the same (or a very similar) distribution. Our training set only partially covers the full space of possible observations, which results in successful avoidance of similarly sized and shaped obstacles, but not in avoidance of very different obstacles since they produce different observations. For moving obstacles, the observations are the same but the underlying true states of the world are different. The training set does not provide any examples of control inputs for moving obstacles.

Due to the nature of neural networks no formal guarantees regarding avoidance or stability can be given. We show experimentally that our approach works well in practice (see Sec. \ref{sec:exp}). Finally, the results presented here are obtained by training the policy on a single static obstacle. The policy can be trained incrementally, e.g. adding larger obstacles.

\subsection{Comparison to related work}
While the proposed learning algorithm bears similarity to DAgger \cite{ross2011reduction}, it differs in important aspects. The proposed \emph{on-policy} step maintains the sample efficiency of the original approach but makes exploration collision-free by using control inputs $\inputVec_{op}$. It has been shown that directly applying outputs from intermediate policies can lead to crashes \cite{KahnZLA16}. We analyze the exploration scheme in depth in Sec.~\ref{sec:results_safety}.

Applying general model based RL, where the true model is obtained during policy training, requires rollouts of the not fully trained policy, which in the case of quadrotors can lead to catastrophic failure \cite{zhang2016learning}. Designing safe model based RL for quadrotors is not a trivial problem and hence adaptive learning techniques based on approximate dynamics have been used \cite{KahnZLA16, zhang2016learning}. We follow this approach.

When tracking the timed reference based on the approximate model using MPC \cite{zhang2016learning}, similar or identical states can be reached at different time steps. This results in ambiguous mappings of \emph{different} control inputs for \emph{similar} or identical states. In the case of MPCC supervision, the control vector $\vuS$ will be the same for a given state $\state_t$. This results in less ambiguous data and a more robust control policy $\policy$ which we experimentally verify in Sec.~\ref{sec:baseline_comparison}.

\section{Experimental Results}\label{sec:exp}
To assess the proposed policy learning scheme we conducted experiments both in simulation (policy trained in simulation) and in real settings (policy trained on real robot). 

\begin{figure}
\centering
\includegraphics[trim={0 0 0pt 0},clip,width=0.95\columnwidth]{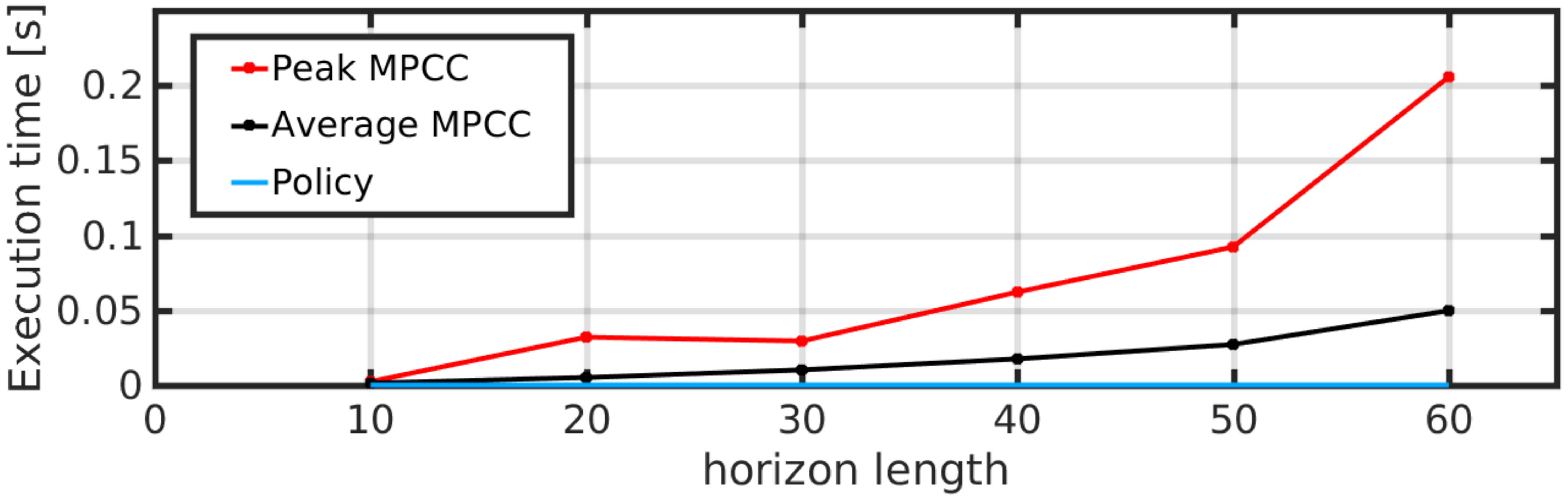}
\caption{\small{\textbf{Execution time:} Horizon length wrt. execution time of controllers. The control policy imitates a long horizon behavior having the same computation time of $2 \cdot 10^{-4}~s$.}}
\label{fig:execution_time}
\end{figure}

\subsection{Implementation Details}

\subsubsection{Global path following}\label{sec:global_path}

The global guidance $\globalPath$ coarsely specifies quadrotor motion, but does not need to be aware of obstacles. The policy controls the $\rollstate, \pitchstate$ angles and the $z$-velocity of the quadrotor, while the yaw angle is controlled separately with a simple PD controller to ensure that the quadrotor always faces the direction of the global path (the distance sensor points in this direction). This parametrization allows for training on straight guidance splines, while at test time this can be applied to arbitrary splines (Fig.~\ref{fig:train_test}, B).

\subsubsection{Hardware and simulation setup}
We evaluate our method in a full physics simulation, using the Rotors quadrotor physics model \cite{rotors} in Gazebo \cite{koenig2004design}, and a Parrot Bebop 2 quadrotor for real world experiments. We use a Vicon system to simulate the sensor readings, using the method described in Sec~\ref{sec:pol_in_out}. In Experiment \ref{sec:mpcc_comparison} we use a simple MATLAB simulation that implements the model given in Sec.~\ref{sec:robot_model}.

\subsection{Comparison with Non-learning Methods }  

\subsubsection{Runtime MPCC vs. policy}
First, we evaluate our method in terms of computational cost by comparing it to a trajectory optimization method. The baseline is a MPCC (cf. Sec. \ref{sec:mpcc}) with an additional collision avoidance cost \cite{NaegeliSIG}. The sampling time of the MPCC is set to $0.1~s$.

Fig.~\ref{fig:execution_time} shows that both average and peak time, measured over 3 iterations, of the MPCC increase depending on the horizon length. The policy can be trained to imitate a long horizon behavior while maintaining constant runtime.

\subsubsection{Policy evaluation - simulation}
\label{sec:eval_policy}
We qualitatively evaluate the learned policy. A long, non-linear guidance is generated and we randomly place obstacles (cf. Fig.~\ref{fig:test_sim}, left). To attain quantitative results we increase the density of the obstacles along the path of a length of $200 m$. For comparison, we use an artificial potential field (APF) method, which has similar computational cost. The potential field pushes the quadrotor to track the global guidance, while repelling it from obstacles. The quadrotor follows a constant velocity reference in the direction of the potential field derivative.

\begin{figure}
\centering
\includegraphics[width=0.99\columnwidth]{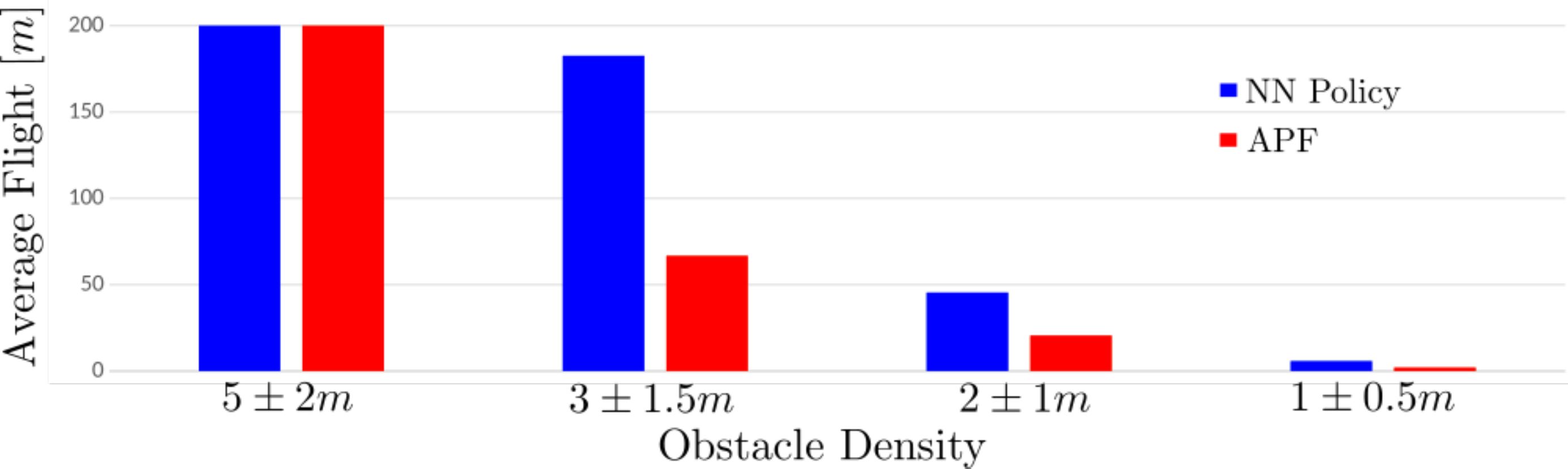}
\caption{\small{\textbf{Average flight distance:} Distance to collision on different obstacle courses (higher is better). Blue (\emph{ours}), red (APF).}}
\label{fig:chart}
\end{figure}

\begin{figure}
\centering
\includegraphics[width=0.99\columnwidth]{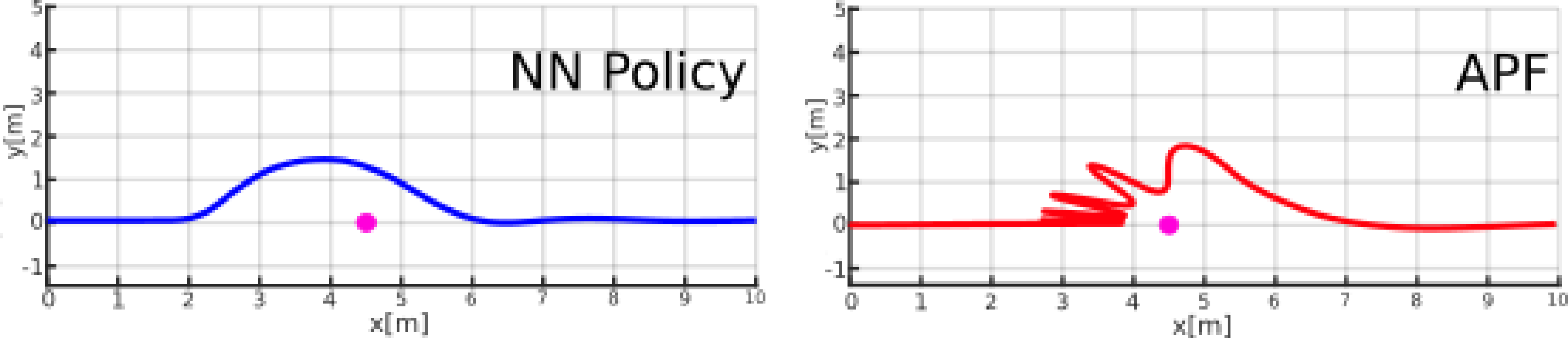}
\caption{\small{\textbf{Comparison trajectories:} Trajectories while avoiding a single obstacle positioned on the guidance $\globalPath$.}}
\label{fig:traj_compare}
\vspace{-12pt}
\end{figure}

\figref{fig:chart} summarizes the average flight distance from three rollouts. The APF velocity reference is set to the average speed of the policy ($1.3 \frac{m}{s}$). For non-trivial cases, the average flight of APF is shorter (cf. Fig. \ref{fig:chart}). Further, the APF method does not consider the robot dynamics which in consequence produces non-smooth trajectories (cf. Fig.~\ref{fig:traj_compare}). Furthermore, APF is only suited for slow maneuvers. Our policy generalizes to much harder cases with obstacles closer to each other than seen at train time. However, once the density surpasses $2 \pm 1 m$ the flight length drops drastically.

\subsection{Supervision Algorithm - Comparison with the Baseline}
\label{sec:baseline_comparison}

One of the main contributions in this work is the MPCC-based path-following supervisor. To evaluate its impact, we compare to a MPC-based trajectory-tracking baseline. For the baseline we obtain an exploration algorithm by augmenting the original MPC objective with the cost in Eq.~(\ref{eq:safe_cost}). Both supervisors are tuned for best learning performance, while producing similar task performance. 

\subsubsection{Single obstacle environment}
\label{sec:mpcc_comparison}

\begin{figure}
\centering
\includegraphics[trim={0pt 0 0pt 0},clip,width=0.9\columnwidth]{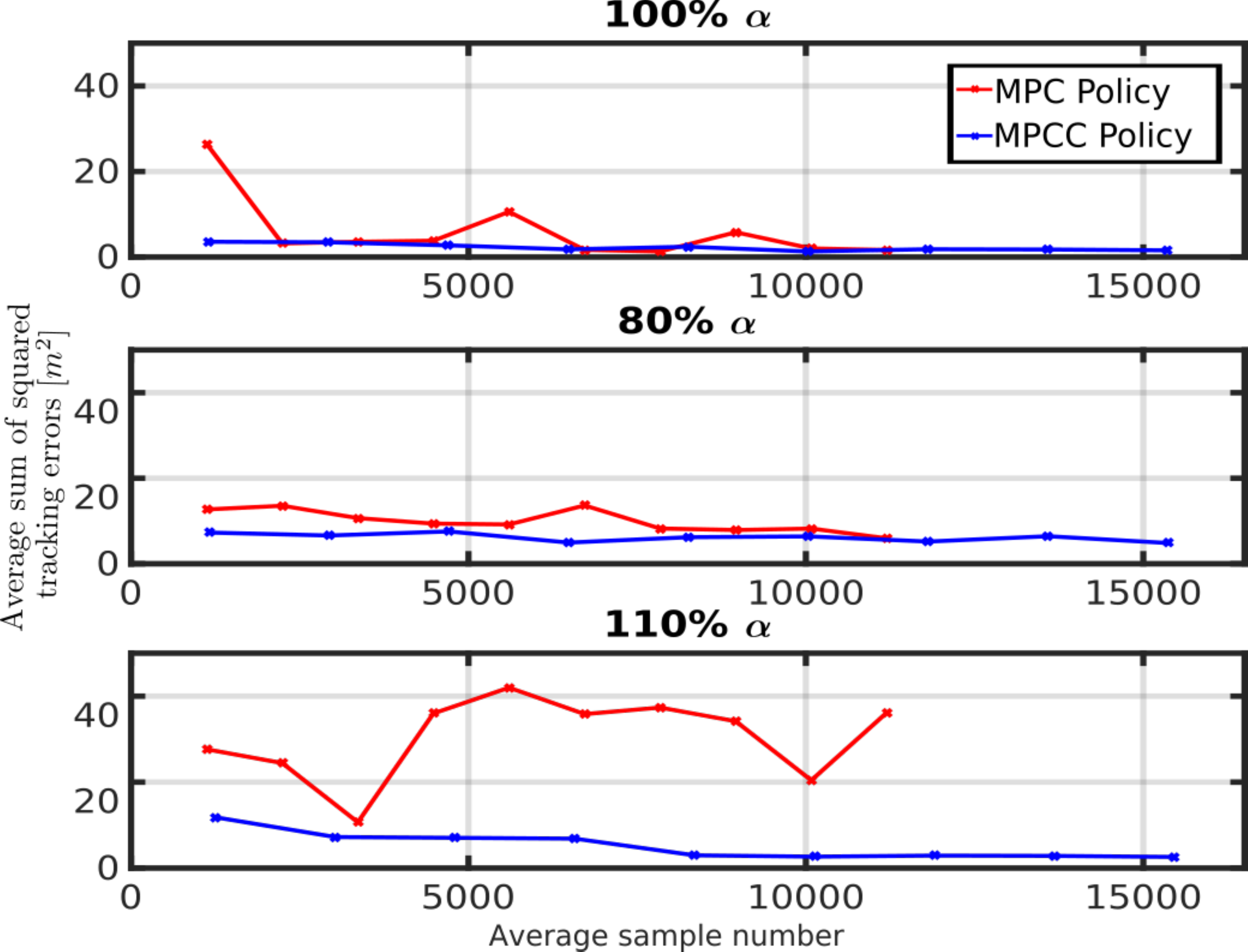}
\caption{\small{\textbf{Policy robustness:} Policy performance as a function of the supervisor. The average error from three trained policies are shown. The error is bounded to $50$. From experiments, we found that error below $10$ gives satisfactory performance. Lower is better.}}
\label{fig:robust}
\end{figure}

To evaluate robustness with respect to model errors we perturb the value of the discretized time constant $\alpha = e^{-\frac{1}{\tau_a}T_s} = 0.85$ used in the supervisor's robot model. Only in this experiment, we use a MATLAB simulation and only use position and velocity measurements as policy inputs.

The task is to learn a single maneuver from four examples, each starting at different positions. At test time we roll-out the policy from six different positions. We compute the error as sum of squared distances of quadrotor positions from the ground truth. This error measures how accurate the policy imitates the supervisor. 

Fig.~\ref{fig:robust} shows that our learning scheme leads to superior robustness and faster convergence compared to the MPC baseline under modeling errors. The baseline achieves desirable scores using the correct model parameters, but convergence behavior is unstable or slower in presence of modeling errors. Even with the true model parameter, MPCC yields faster convergence behavior. Besides modeling errors we have unmodeled effects on the real system, which may lead to unstable convergence of MPC-based schemes.

\subsubsection{Multi-obstacle environment}

In this experiment we compare the MPCC supervisor to the MPC baseline in the Gazebo simulator. The simulator implements complex quadrotor dynamics \cite{rotors}. Contrary to the previous experiment, the policies are trained for the final task i.e. guidance tracking with collision avoidance. We train the policies on the same number of examples (12). The examples for the MPCC supervisor are generated by our algorithm, while the examples for MPC are generated by the off-line trajectory optimization algorithm. The trajectory optimization cost is adjusted so that the quadrotor follows the global guidance with constant speed while avoiding the obstacles.  

Table \ref{tbl:policies_compare} summarizes the results. We were able to train the policy with a MPC supervisor, but the performance of the policy was not satisfactory. The first issue is that the quadrotor cannot follow the global guidance, drifting from the prescribed path in the $z$ direction. Although the policy performance is not satisfactory, we still evaluated the policy on the obstacle course. On the obstacle course with density of $3 \pm 1.5~m$ along the path, the average flight length of the MPC policy is $41.67 ~ m$ which is significantly lower than the policy trained with the MPCC ($183.3 ~ m$) supervisor. In the light of the previous experiments these results are logical since the policy trained with MPC is not able to accurately follow trajectories in the presence of modeling errors. 

\begin{table}
\caption{Comparison with the Baseline Supervisor}
\begin{center}
\begin{tabular}{| p{3.5cm} | l | l |}
\hline
Task & MPC policy & MPCC policy \\
\hline
Max. tracking deviation $z$ axis & $0.847~m$ & $0.077~m$ \\
\hline
Average flight length & $41.67~m$ & $183.3~m$ \\
\hline
\end{tabular}\label{tbl:policies_compare}
\end{center}
\vspace{-12pt}
\end{table}

\subsection{Evaluation of Collision-free Exploration}
\label{sec:results_safety}
The choice of the contouring penalty directly impacts which states are being visited in the on-policy step. Table \ref{tbl:contour_penalty} summarizes results for different values, measured as sum of squared distances from the example paths during training and from ground truth at test time under different parameters (averaged over trajectories).

\begin{table}
\caption{Exploration algorithm performance}
\begin{center}
\begin{tabular}{| p{1.5cm} | p{1.1cm} | l | l | l |}
\hline
        & \emph{unsafe}  & $K_c = 0.1$ & $K_c = 10$ & $K_c = 25$\\
\hline
Collisions & Train~time           & Test time         & No         & Test time \\
\hline
Train error &  /        & $2.75~m$ & $1.38~m$ & $1.02~m$  \\
\hline
Test error &  /          & $1.99~m$ & $0.84~m$ & $2.55~m$  \\
\hline
\end{tabular}\label{tbl:contour_penalty}
\end{center}
\end{table}

We were not able to train the policy by using intermediate policies for exploration in the on-line step of the algorithm (\emph{unsafe}). A too small contouring cost ($K_c = 0.1$) leads to large deviation from the example path (high train error) and results in poor generalization (high test error). Too large penalization of the contouring cost ($K_c = 25.0$) suppresses exploration and leads to overfitting (high test error).

\begin{figure}
\centering
\includegraphics[width=0.99\columnwidth]{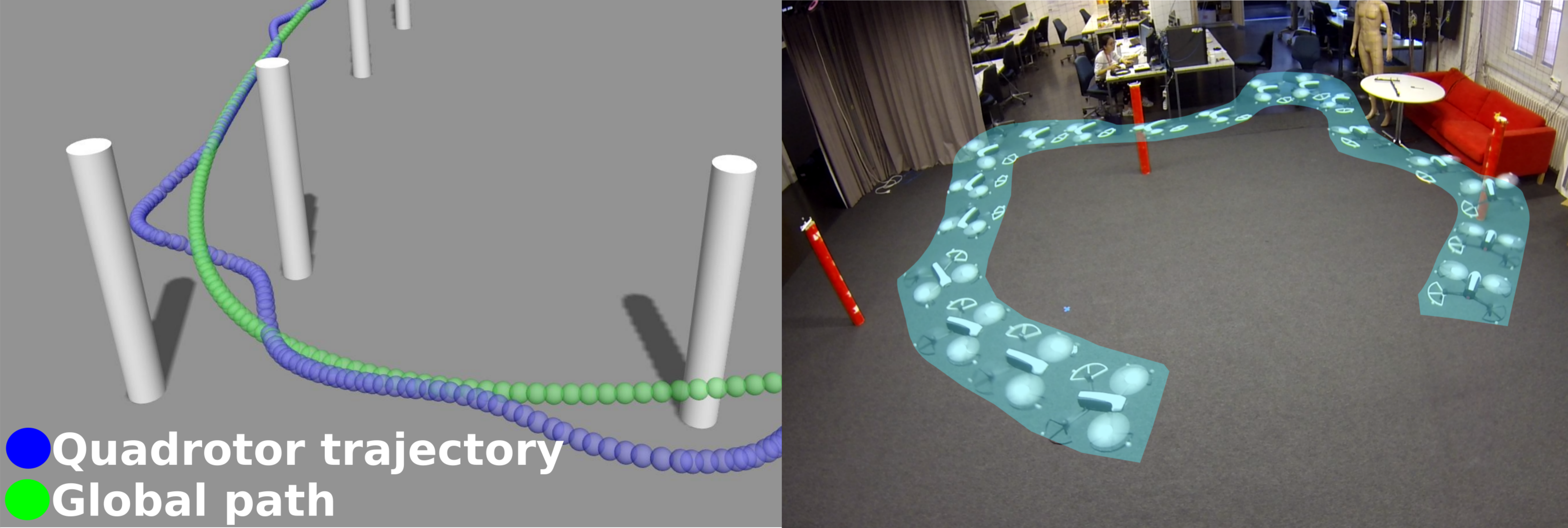}
\caption{\small{Left: \textbf{Policy roll-out:} Unseen test scene including long guidance (green), obstacles and flown policy roll-out (blue). Right: \textbf{Static obstacle.} Policy roll-out in the real environment. Three obstacles are positioned along a circular reference.}}
\label{fig:test_sim}
\vspace{-12pt}
\end{figure}

\subsection{Policy Evaluation}

\subsubsection{Policy generalization - simulation}

We test the generalization ability with different obstacles. Various courses are obtained as in Sec. \ref{sec:eval_policy} (density $3 \pm 1.5$), and we change the obstacle types. We increase the obstacle radius up to $50~\%$ where the policy begins to predict invalid outputs (NaN). The policy successfully avoids cubic obstacles of similar size as the training obstacles. We conclude that the size of the obstacle is the critical factor for generalization.

Next, we evaluate the policy on obstacles that are moving perpendicular to the global guidance path. The obstacle velocity is gradually increased, until collision occurs at $0.7\frac{m}{s}$. Moving obstacles reduce the effective lateral robot speed and no such behavior was observed during training.

\subsubsection{Policy evaluation - real} We conduct similar experiments on a physical quadrotor, positioning obstacles directly on the desired path (Fig.~\ref{fig:test_sim}, right). Due to the small experimental space we reduce the avoidance onset to $d=2 m$. No collisions occur and the course is always completed.

A final experiment evaluates policy under moving obstacles such as humans. In our experiments the robot successfully avoids slow moving targets, keeping away from the human at distances similar to training time. Please refer to the  video ( https://youtu.be/eEqzhglPjNE ) for more results.

\section{Conclusion}
We have proposed a method for learning control policies using neural networks in imitation learning settings. The approach leverages a \emph{time-free} MPCC path following controller as a supervisor in both off-policy and on-policy learning.
We experimentally verified that the approach converges to stable policies which can be rolled out successfully to unseen environments both in simulation and in the real-world.
Furthermore, we demonstrated that the policies generalize well to unseen environments and have initially explored the possibility to roll out policies in dynamic environments.

\bibliographystyle{IEEEtran}

\bibliography{IEEEabrv,references}

\begin{thebibliography}{10}
\providecommand{\url}[1]{#1}
\csname url@rmstyle\endcsname
\providecommand{\newblock}{\relax}
\providecommand{\bibinfo}[2]{#2}
\providecommand\BIBentrySTDinterwordspacing{\spaceskip=0pt\relax}
\providecommand\BIBentryALTinterwordstretchfactor{4}
\providecommand\BIBentryALTinterwordspacing{\spaceskip=\fontdimen2\font plus
\BIBentryALTinterwordstretchfactor\fontdimen3\font minus
  \fontdimen4\font\relax}
\providecommand\BIBforeignlanguage[2]{{%
\expandafter\ifx\csname l@#1\endcsname\relax
\typeout{** WARNING: IEEEtran.bst: No hyphenation pattern has been}%
\typeout{** loaded for the language `#1'. Using the pattern for}%
\typeout{** the default language instead.}%
\else
\language=\csname l@#1\endcsname
\fi
#2}}

\bibitem{grzonka2012fully}
S.~Grzonka, G.~Grisetti, and W.~Burgard, ``A fully autonomous indoor
  quadrotor,'' \emph{IEEE Transactions on Robotics}, vol.~28, pp. 90--100,
  2012.

\bibitem{mueller2013model}
M.~W. Mueller and R.~D'Andrea, ``A model predictive controller for quadrocopter
  state interception,'' \emph{Control Conference (ECC), 2013 European}, pp.
  1383--1389, 2013.

\bibitem{oleynikova2016continuous}
H.~Oleynikova, M.~Burri, Z.~Taylor, J.~Nieto, R.~Siegwart, and E.~Galceran,
  ``Continuous-time trajectory optimization for online uav replanning,'' in
  \emph{Intelligent Robots and Systems (IROS), 2016 IEEE/RSJ International
  Conference on}.\hskip 1em plus 0.5em minus 0.4em\relax IEEE, 2016, pp.
  5332--5339.

\bibitem{pivtoraiko2013incremental}
M.~Pivtoraiko, D.~Mellinger, and V.~Kumar, ``Incremental micro-uav motion
  replanning for exploring unknown environments,'' in \emph{Robotics and
  Automation (ICRA), 2013 IEEE International Conference on}.\hskip 1em plus
  0.5em minus 0.4em\relax IEEE, 2013, pp. 2452--2458.

\bibitem{frew2004obstacle}
E.~Frew and R.~Sengupta, ``Obstacle avoidance with sensor uncertainty for small
  unmanned aircraft,'' in \emph{Decision and Control, 2004. CDC. 43rd IEEE
  Conference on}, vol.~1.\hskip 1em plus 0.5em minus 0.4em\relax IEEE, 2004,
  pp. 614--619.

\bibitem{rodriguez2014trajectory}
E.~J. Rodr{\'\i}guez-Seda, C.~Tang, M.~W. Spong, and D.~M. Stipanovi{\'c},
  ``Trajectory tracking with collision avoidance for nonholonomic vehicles with
  acceleration constraints and limited sensing,'' \emph{The International
  Journal of Robotics Research}, vol.~33, no.~12, pp. 1569--1592, 2014.

\bibitem{faust2014automated}
A.~Faust, I.~Palunko, P.~Cruz, R.~Fierro, and L.~Tapia, ``Automated aerial
  suspended cargo delivery through reinforcement learning,'' \emph{Artificial
  Intelligence}, 2014.

\bibitem{mnih2013playing}
V.~Mnih, K.~Kavukcuoglu, D.~Silver, A.~Graves, I.~Antonoglou, D.~Wierstra, and
  M.~Riedmiller, ``Playing atari with deep reinforcement learning,''
  \emph{arXiv preprint arXiv:1312.5602}, 2013.

\bibitem{deisenroth2013survey}
M.~P. Deisenroth, G.~Neumann, J.~Peters, \emph{et~al.}, ``A survey on policy
  search for robotics,'' \emph{Foundations and Trends{\textregistered} in
  Robotics}, vol.~2, no. 1--2, pp. 1--142, 2013.

\bibitem{abbeel2007application}
P.~Abbeel, A.~Coates, M.~Quigley, and A.~Y. Ng, ``An application of
  reinforcement learning to aerobatic helicopter flight,'' in \emph{NIPS},
  2007.

\bibitem{levine2016end}
S.~Levine, C.~Finn, T.~Darrell, and P.~Abbeel, ``End-to-end training of deep
  visuomotor policies,'' \emph{Journal of Machine Learning Research}, vol.~17,
  no.~39, pp. 1--40, 2016.

\bibitem{KahnZLA16}
G.~Kahn, T.~Zhang, S.~Levine, and P.~Abbeel, ``Plato: Policy learning using
  adaptive trajectory optimization,'' \emph{Robotics and Automation (ICRA),
  2017 IEEE International Conference on}, pp. 3342--3349, 2017.

\bibitem{ross2011reduction}
S.~Ross, G.~J. Gordon, and D.~Bagnell, ``A reduction of imitation learning and
  structured prediction to no-regret online learning.'' in \emph{AISTATS},
  vol.~1, no.~2, 2011, p.~6.

\bibitem{ross2013learning}
S.~Ross, N.~Melik-Barkhudarov, K.~S. Shankar, A.~Wendel, D.~Dey, J.~A. Bagnell,
  and M.~Hebert, ``Learning monocular reactive uav control in cluttered natural
  environments,'' in \emph{Robotics and Automation (ICRA), 2013 IEEE
  International Conference on}.\hskip 1em plus 0.5em minus 0.4em\relax IEEE,
  2013, pp. 1765--1772.

\bibitem{mordatch2015interactive}
I.~Mordatch, K.~Lowrey, G.~Andrew, Z.~Popovic, and E.~V. Todorov, ``Interactive
  control of diverse complex characters with neural networks,'' in \emph{NIPS},
  2015.

\bibitem{zhang2016learning}
T.~Zhang, G.~Kahn, S.~Levine, and P.~Abbeel, ``Learning deep control policies
  for autonomous aerial vehicles with mpc-guided policy search,'' in
  \emph{Robotics and Automation (ICRA), 2016 IEEE International Conference
  on}.\hskip 1em plus 0.5em minus 0.4em\relax IEEE, 2016, pp. 528--535.

\bibitem{lam2010model}
D.~Lam, C.~Manzie, and M.~Good, ``Model predictive contouring control,'' in
  \emph{49th IEEE Conference on Decision and Control (CDC)}.\hskip 1em plus
  0.5em minus 0.4em\relax IEEE, 2010, pp. 6137--6142.

\bibitem{NaegeliSIG}
T.~N{\"a}geli, L.~Meier, A.~Domahidi, J.~Alonso-Mora, and O.~Hilliges,
  ``Real-time planning for automated multi-view drone cinematography,'' in
  \emph{ACM Transactions on Graphics (Proceedings of SIGGRAPH)}, 2017.

\bibitem{forcesproweb}
A.~Domahidi and J.~Jerez, ``{FORCES Pro: code generation for embedded
  optimization},'' 2016, https://www.embotech.com/FORCES-Pro.

\bibitem{rotors}
F.~Furrer, M.~Burri, M.~Achtelik, and R.~Siegwart, ``Robot operating system
  (ros),'' \emph{Studies Comp.Intelligence Volume Number:625}, vol. The
  Complete Reference (Volume 1), p. Chapter 23, 2016.

\bibitem{koenig2004design}
N.~Koenig and A.~Howard, ``Design and use paradigms for gazebo, an open-source
  multi-robot simulator,'' in \emph{Intelligent Robots and Systems (IROS), 2004
  IEEE/RSJ International Conference on}, vol.~3.\hskip 1em plus 0.5em minus
  0.4em\relax IEEE, 2004, pp. 2149--2154.

\end{thebibliography}

\end{document}